\documentclass[11pt,a4paper]{article}
\usepackage{authblk}

\usepackage[hyperref]{acl2019}
\usepackage{balance}
\usepackage{times}
\usepackage{latexsym}
\usepackage{graphicx}
\usepackage{url}
\usepackage{todonotes}

\usepackage{tikz}
\usepackage{booktabs}
\usepackage{xcolor}
\usepackage{amsmath}
\usepackage{comment}
\usepackage{enumitem}
\usepackage{paralist} 
\usepackage{multicol}

\title{Time-Out: Temporal Referencing for Robust Modeling of\\ Lexical Semantic Change}

\author{Haim Dubossarsky$^{\spadesuit}$~~~Simon Hengchen$^{\diamondsuit}$~~~Nina Tahmasebi$^{	\clubsuit}$~~~Dominik Schlechtweg$^{\heartsuit}$ \thanks{ ~ The order has been randomly determined and all authors contributed equally to this work. }\\

$^{\spadesuit}$ Language Technology Lab, University of Cambridge\\
$^{\diamondsuit}$ COMHIS, University of Helsinki \\
$^{\clubsuit}$ Department of Swedish, University of Gothenburg \\
$^{\heartsuit}$ Institute for Natural Language Processing, University of Stuttgart \\
{\tt hd423@cam.ac.uk}~~~{\tt simon.hengchen@helsinki.fi} \\ {\tt nina.tahmasebi@gu.se}~~~{\tt schlecdk@ims.uni-stuttgart.de}}

\date{March 2019}

\aclfinalcopy

\begin{document}

\maketitle

\begin{abstract}
State-of-the-art models of lexical semantic change 
detection suffer from noise stemming from vector space alignment. We  have  empirically  tested  the \textit{Temporal Referencing} method  for  lexical  semantic  change and show that, by avoiding alignment, it is less affected by this noise. We show that, trained on a diachronic corpus, the skip-gram with negative sampling architecture with temporal referencing  outperforms alignment models on a synthetic task as well as a manual testset. We introduce a principled way to simulate lexical semantic change and systematically control for possible biases.
\end{abstract}

\section{Introduction}

These past years have seen the rise of computational methods to detect, track, qualify, and quantify how a word's sense -- or senses -- change over time.
These tasks are critical challenges that are relevant to a range of NLP fields, including the study of 
historical semantic change. The successful outcome of semantic change detection is relevant to any diachronic textual analysis, including machine translation or normalization of historical texts \citep{BollmannEtal17}, the detection of cultural semantic shifts \citep{kutuzov-etal-2017-tracing} or applications in digital humanities \citep{Tahmasebi17}.
However, currently, the best-performing models  \citep{DiachronicWordEmb,kulkarni2015statistically,Schlechtwegetal19} require a complex alignment procedure and have been shown to suffer from biases \citep{Dubossarsky-EMNLP17}. This exposes them to various sources of noise influencing their predictions; a fact which has long gone unnoticed because of the lack of standard evaluation procedures in the field. 

We examine the modeling approach of \textit{Temporal Referencing} (TR) which avoids post hoc alignment and is applicable to any vector space learning technique.  We show that it (i) is less affected by noise and (ii) clearly outperforms state-of-the-art alignment models on a synthetic change detection task. The task is based on data from a synchronic corpus into which we artificially inject lexical semantic change (LSC) in a controlled and semantically principled way. We further evaluate the models on a manual testset of diachronic LSC and examine their properties. 

In this paper, we focus on skip-gram with negative sampling (SGNS) models \cite{Mikolov13b} and PPMI \cite{Levy2015}  and make use of TR to share context information across time periods, while learning individual embeddings for a target word in each time period. We evaluate models in two ways: on the one hand, through the comparison of model performance between semantically changing and stable words. This is achieved through the synthetic introduction (and removal) of polysemy, mimicking \citet{schutze-1998,kulkarni2015statistically,rosenfeld2018deep}. We differ from previous work by creating those changes in a more structured way, and for many time points. The second type of evaluation put forward is a study built on a smaller number of words manually classified as changed or stable.

Our contributions are the following:
\vspace{0.1cm}
\begin{compactitem}
\item \textbf{Noise Reduction:} We avoid post hoc alignment by TR and show that it outperforms other models and is robust to noise.
\item \textbf{LSC Simulation:} We propose a systematic and principled method of injecting semantic change in a controlled fashion. 
\item \textbf{Evaluation:} We evaluate (i) by testing for noise reduction in a control condition, (ii) on large and controlled artificial data and (iii) on a manually annotated LSC testset. 
\item \textbf{Framework:} The above comprises a framework to test \textit{any} model of semantic change for their levels of noise and sensitivity in detecting simulated semantic change.
\end{compactitem}

\section{Related Work}
\label{sec:relatedwork}

\paragraph{Models of LSC Detection} Computational approaches to semantic change detection can be divided in different families:  count-based semantic spaces \citep{sagi2009semantic,Gulordava} and more recently based on neural embeddings \citep{TempAnal-NeuralLangModel, basilediachronic, kulkarni2015statistically,DiachronicWordEmb}; graph-based models \citep{Tahmasebi17, WSE-Mitra,Mitra2015NLE}; and finally topic-based \cite{NovelSenses,
wang2015sense,frermann2016bayesian,hengchen2017does,perrone2019gasc}. 
Recently, we have seen dynamic embeddings with the main aim to circumvent alignment, and share data across time points, thus reducing data volume requirements. Using different base embeddings, SGNS \cite{bamler17}, PPMI \cite{Yao-2018}, and Bernoulli embeddings \cite{RudolphB18-dynamicEmbforLangEvo}, the results show that sharing data is beneficial regardless of the method.\footnote{For an extensive survey of computational approaches to lexical semantic change, we refer the readers to \citet{tahmasebi2018survey}, and to \citet{kutuzov-etal-2018} for a specialized focus on diachronic word embeddings.} Temporal Referencing has been applied first in the field of term extraction \citet{ferrari2017detecting} and recently been tested for diachronic LSC detection \cite{Schlechtwegetal19}.

\paragraph{Evaluation} Due to a lack of proper evaluation methods and datasets, all papers above have performed different, non-comparable evaluations. 
Previous evaluation procedures mainly tackle a few words: case studies of individual words \citep{Wijaya,Jatowt:2014:FAS:2740769.2740809,hamilton-etal-2016}, or a comparison between a few changing and semantically stable words \citep{NovelSenses,schlechtweg-EtAl:2017:CoNLL}. Other works focus on the post hoc evaluation of their respective models \citep{kulkarni2015statistically,Eger-SemChange}. Importantly, \citet{Dubossarsky-EMNLP17} proposed to use a control condition to mitigate the absent of validated evaluation methods and datasets.

\paragraph{Control Condition}\label{sec:controlcondition}
Evaluating empirical results often demands comparing these under a control condition in order to maintain that these are indeed valid and are not the result of unwanted confounding factors. A control condition directly follows from a specific research hypothesis, and therefore must resemble the original condition in any aspect, except the variable of interest that is being hypothesized about. For example, \citet{Dubossarsky-EMNLP17} attested that a shuffled diachronic corpus is a proper control condition to test models for semantic change, under the hypothesis that such models indeed capture semantic change and not something else. They concluded that any degree of semantic change that is reported by a model on the shuffled corpus may only be related to noise, instead of a true semantic change. Similarly, we propose to test the noise levels associated with different semantic change models using a shuffled historical corpus, and evaluate their true degree of semantic change by comparing their results to the original historical corpus. Importantly, there are many ways to create control conditions, and the synthetic lexical semantic change proposed in Section \ref{sec:SynLSC} contains another type of control condition, that is based on artificially induced semantic change.

\section{Models}
\label{sec:models}

\paragraph{Embeddings} A common method in LSC detection is to learn low-dimensional semantic vector spaces (embeddings) for specific time periods and then align spaces for consecutive time periods with an orthogonal mapping which minimizes the distances between the time-specific vectors for all words \citep{DiachronicWordEmb}. Given two consecutive time periods $a$, $b$, and corresponding text corpora $C_a$, $C_b$, we learn two vector spaces $A$, $B$. Orthogonal Procrustes analysis can then be applied to find the optimal mapping matrix $W^{*}$ such that the sum of squared Euclidean distances between $B$'s mapping $BW$ and $A$ is minimized:
\begin{equation*}
W^{*} = \arg\min_{W} \| BW-A \|^{2}.
\end{equation*}
The optimal solution for this problem is given by an application of Singular Value Decomposition \citep{artetxe2017acl}.\footnote{$W$ is constrained to be orthogonal. $A$ and $B$ are first length-normalized and mean-centered and their rows are reduced to the intersection of the vocabulary of $C_a$ and $C_b$ for finding the mapping.} The degree of LSC of a word $w$ is then measured with the cosine distance \citep{salton1986introduction} between $w$'s vectors in $A$ and $BW^{*}$ ($B$'s mapping). This approach has been found to outperform other LSC detection methods in various studies \citep{DiachronicWordEmb,kulkarni2015statistically}. It has the advantage of not assuming that words keep the same meaning over time. A presumable downside of this approach is expected noise from the alignment, i.e., it may not be possible to align all words to each other that have similar meanings, because the spaces were learned independently.

\paragraph{PPMI} Another method to learn time-specific semantic vector space representations $A$, $B$ is to store count-based co-occurrence information for each word in a high-dimensional sparse matrix and then apply Positive Pointwise Mutual Information (PPMI) weighting \cite{Levy2015}. In such a matrix each column stores the co-occurrence statistics with a specific context word. This has the advantage that $A$ and $B$ can be aligned straightforwardly, because many context words occur as columns in both $A$ and $B$ and can hence be mapped onto each other. Mapping $A$ and $B$ to a common coordinate axis then corresponds to intersecting their columns \citep{DiachronicWordEmb}. This has the advantage of avoiding the complex alignment procedure for embeddings, but also loses their performance advantages \cite{marcobaroni2014predict,Levy2015}.

\paragraph{Temporal Referencing}\label{sec:tr}
Temporal Referencing (TR) is an alternative to learning individual word representations for different time periods, which avoids alignment using a procedure radically simpler than proposed for dynamic embeddings. TR is potentially applicable to every vector space learning method.  We treat all time-specific corpora $C_a$, $C_b$, ..., $C_n$ as one corpus $C$ and learn word representations on the full corpus. However, we first replace each target word $w\in C_t$ with a time-specific token $w_{t}$.\footnote{In our case, $t$ is a decade. E.g., in the corpus for 1920 we replace each occurrence of \textit{computer} with the string \textit{computer$_{1920}$}.}
This temporal referencing of $w$ is only performed when it is a target word, when the word is considered a context word, it remains unchanged. Following this procedure, we learn one single space that contains a vector for each target-time pair $w_{t}$, which may  be compared directly without the need for alignment.
Besides the considerable advantages of avoiding alignment and being applicable to count-based and embedding methods, it presumably lowers data requirements (because context words are collapsed, and thus shared, across corpora). Accordingly, we assume TR to produce smoother change values. As various other models, TR relies on the assumption that the semantics of the context words stays relatively stable over time.

\section{Synthetic Lexical Semantic Change}
\label{sec:SynLSC}

\begin{figure}[t]
        \center{\includegraphics[width=1.0\linewidth]  {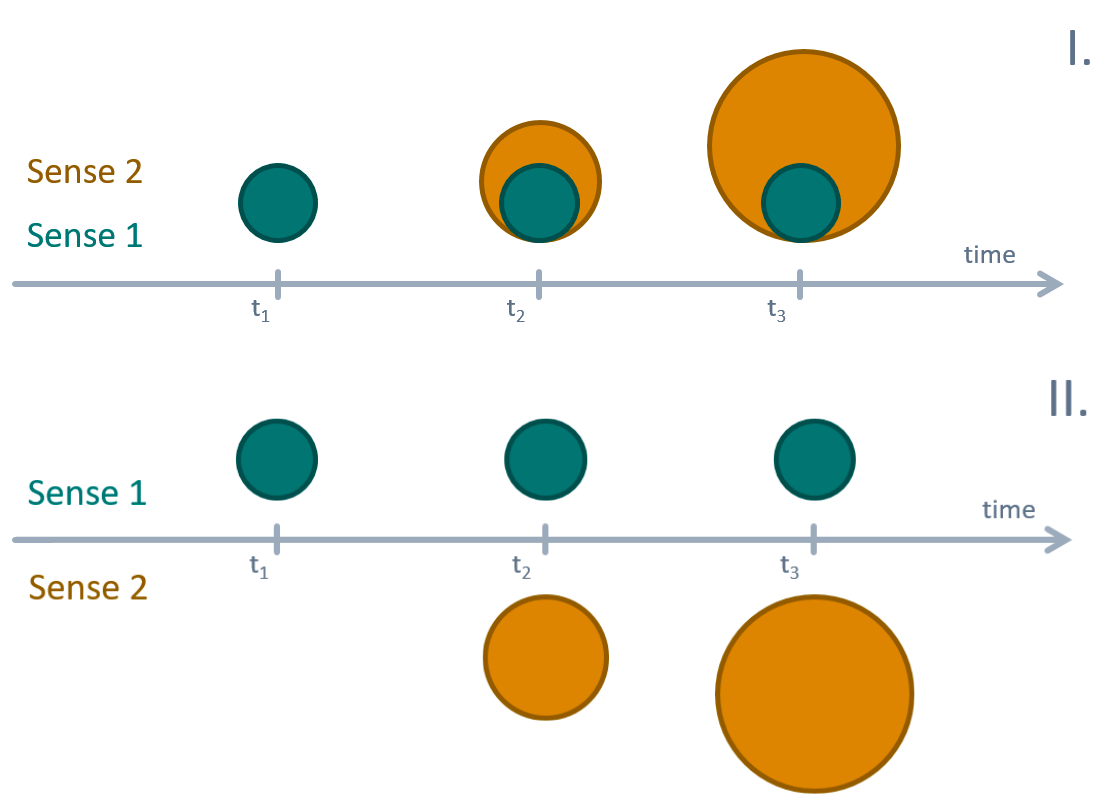}} %{bothChange_quality.png}}
        \caption{\label{fig:synChange}  Increase in semantic material for a word by means of sense injection. I.: new injected sense is related to the existing sense. II.: new injected sense is unrelated.}
      \end{figure}

We aim to simulate semantic change under controlled settings, while keeping the corpus as natural as possible.\footnote{Hence, the target words' frequencies were not matched, but rather stayed natural.} We call this procedure \textit{sense injection}. We increase the semantic material of a recipient word $w^r$ in subsequent subcorpora by injecting contexts from a donor word $w^d$. The context of the recipient word (illustrated as Sense 1 in Figure \ref{fig:synChange}) stays as it is in the corpus. The first subcorpus contains only contexts from the recipient $w^r$ and all the contexts of the donor $w^d$ are removed. In the next time period we  add 25\% of the contexts of $w^d$, with donor word replaced by the recipient word. In each subsequent corpus, an additional 25\% of the donor word are injected until the last time periods contain equal amounts of contexts from the donor and recipient. As a result, seen from the recipient $w^r$, the last time periods have double the amount of contexts as in the first time period $|w^r({t_n}) +w^d({t_n})| = 2*|w^r({t_1})|$. 

Note that due to the polysemous nature of words (each is usually associated with more than one sense), we preferred to \textit{add} the donor words' contexts instead of simply \textit{replacing} the existing contexts of the recipient words with the contexts of the donor words. This is because the former involves a single source of synthetic lexical semantic change, while the latter involves two sources (the removal of contexts associated with different senses of a recipient word, as well as the added contexts associated with the senses of a donor word). As a result, this procedure yields less noisy examples of synthetic lexical semantic change.

We differ between cases where recipient and donor are related (e.g. \textit{maker} $\rightarrow$ \textit{creator}, Fig. \ref{fig:synChange}a) and unrelated (e.g. \textit{shoulders} $\rightarrow$ \textit{horde}, Fig. \ref{fig:synChange}b), following e.g., \citet{pilehvar2013paving}. This procedure is aimed to give us insight into how much novel semantic material is needed for our methods to detect semantic change. Our hypothesis is that cases where the donor word is unrelated to the recipient word should be simpler to detect compared to those that are in close relation. It is linguistically motivated to choose semantically related words to simulate sense change; those are the most difficult cases of sense change, and a likely procedure of semantic change introducing polysemy \cite{Blank97XVI}. 
 
Finally, to simulate the same increase in frequency, we repeat the sense injection for a set of control words. In this case recipient and donor word are the same $w^r = w^d$. This creates the same increased frequency of the recipient word $|w^r({t_n})| = 2*|w^r({t_1})|$ as the above, but without any added semantic information because the control word keeps its original contexts.

%%% - same colour small and big orange
%%% - Sense 2 in dark orange
%%% - sense key in larger font 

\section{Experimental setup}
\subsection{Corpora}\label{sec:corpora}
For Experiment 1 (Sec.~\ref{subsec:experiment1}) we used COHA \citep{davies2002corpus}, of which we restrict ourselves to decadal bins spanning from 1920 to 1970 so as to have a comparable number of tokens for each time slice. 
For Experiment 2 (Sec.~\ref{subsec:experiment2}) we used COCA \citep{davies2008corpus}, of which we remove the spoken and academic genres in order to maintain a more similar usage context of words.
As a control setting, we created shuffled versions of the same corpora with the same periods, and straightforwardly followed \citet{Dubossarsky-EMNLP17}.

\subsection{Synthetic semantic change}
\label{Synthetic semantic change}
For related words, we used the Noun-Noun pairs in SimLex-999 \cite{hill2015simlex} as a starting point. However, even semantically unrelated pairs in SimLex were deemed somewhat related by our annotators, and therefore we kept only 10 of those. We created the rest of the list of unrelated words as follows:
we randomly sampled 300 lowercased nouns\footnote{The filtering was carried out on the basis of the output of NLTK \citep{bird2009natural}'s $pos\_tag()$ function.} from our corpus, which we assembled into 150 pairs. 
We then asked three annotators to independently go through the list of generated pairs and determine whether they were semantically related or not.
All 150 pairs were deemed semantically unrelated by at least 2 annotators. Only 5 pairs had a disagreement but were qualified as border line cases by the disagreeing annotator, and kept. 
This procedure yielded 356 word pairs in total, of which 196 were related and 160 were not related.

\subsection{Model training}
We tested two models in our experiments: (i) low-dimensional embeddings learned with SGNS and (ii) high-dimensional sparse PPMI vectors. Each of these were tested with their respective alignment method (AL) and with Temporal Referencing (TR) as described in Section \ref{sec:models}, leaving us with four models to compare:
    \begin{center}
    \begin{tabular}{ll}
      SGNS$_{\text{AL}}$   &  SGNS$_{\text{TR}}$\\
       PPMI$_{\text{AL}}$  & PPMI$_{\text{TR}}$\\
    \end{tabular}
    \label{tab:my_label}
    \end{center}
In order to avoid that replaced target words co-occur with other target words in TR we used the implementation of \citet{Levy2015}, allowing us to train SGNS and PPMI on extracted word-context pairs instead of the corpus directly. For this, we iterated over corpus $C_t$ such that for each token $w$ and for each of its context words $c$ within a symmetric window we extracted the word-context pair:
($w_t$,$c$) if $w$ is a target word and ($w$,$c$) otherwise.

In this way, we guarantee a target word is never replaced and treated as context of any other word. For TR, SGNS and PPMI were then trained on these extracted pairs. For AL, we extracted only regular word-context pairs ($w$,$c$) and trained SGNS and PPMI on these. LSC is measured for all four models via cosine distance.\footnote{Find a full implementation of the pipeline at \url{https://github.com/Garrafao/TemporalReferencing}.} (See Appendix \ref{sec:parameter} for preprocessing and hyper-parameter details.)

\section{Evaluation}
% is it clear what the two control conditions are?!
To test our methods we performed three main experiments, comparing the performances of TR to the existing state-of-the-art diachronic model alignment.
In the first experiment, we compare the models' performance under control conditions that address complementary (potential) weaknesses.
The second experiment tests different synthetic change types and assesses whether better models improve detection of lexical semantic change, in a controlled setting.
Finally, we test our methods on a manually created testset on a genuine corpus, and manually inspect the results. 

\subsection{Experiment 1: Model comparison}
\label{subsec:experiment1}

In this experiment, we trained each model on two corpora, one genuine diachronic corpus with natural semantic change, and one shuffled where the diachronic change is distributed equally across all time periods (see Sec. \ref{sec:corpora}).  We study the average change of cosine distance as a proxy for semantic change. Following \citet{Dubossarsky-EMNLP17} we consider the average cosine distance (\textit{\textbf{acd}}) trained on the genuine corpus to correspond to true semantic change + noise. In contrast, the average cosine distance on the shuffled corpus corresponds to pure noise. Therefore, the difference between the two equals to true signal, or in other words, true lexical semantic change. 

Importantly, we are interested in investigating, and hopefully mitigating, possible sources of the noise that might be found in some of the models. Specifically, we hypothesize that the alignment procedure adds considerable noise to the \textit{acd}, and plan to test how TR can alleviate some of that noise. Moreover, TR is assumed to contribute not only by circumventing the alignment, but also by producing more stable context vectors due to the increased amount of data on which they are trained.\footnote{We differ between \textit{stable} vectors that do not change despite the randomness involved in training between multiple runs, and \textit{accurate} vectors give a good representation of meaning. Note that when we use the term \textit{stable word} we mean stable in meaning over time.} Therefore, we first tease-apart these factors using the following comparisons between the different models.
\vspace{0.1em}
\begin{compactenum}
\item For all models, we consider the difference in average cosine distance between genuine and shuffled conditions ($acd_{genuine} -acd_{shuffled}$) as being inversely proportional to the amount of noise that the original model unknowingly captures. Hence, the larger the difference, the less noisier (and better) the model is. We consider this to be an approximation of the \textit{true semantic change}.
\item Focusing on the differences between the two PPMI models allows us to test the independent contribution of TR in providing more accurate context vectors because the intersection of the PPMI vectors are inherently aligned.
\item Focusing on the SGNS models conflates the potential benefits from more accurate context vectors with the disadvantage of Procrustes alignment (which is necessary for SGNS$_{\text{AL}}$ but not for SGNS$_{\text{TR}}$).
\item The difference between the last two would allow us to evaluate the independent contribution of these two sources on the (presumably) less noisy SGNS$_{\text{TR}}$ model scores.
\end{compactenum}

\paragraph{Results (experiment 1)}

\begin{figure}[h]
\center{\includegraphics[scale=0.5]        {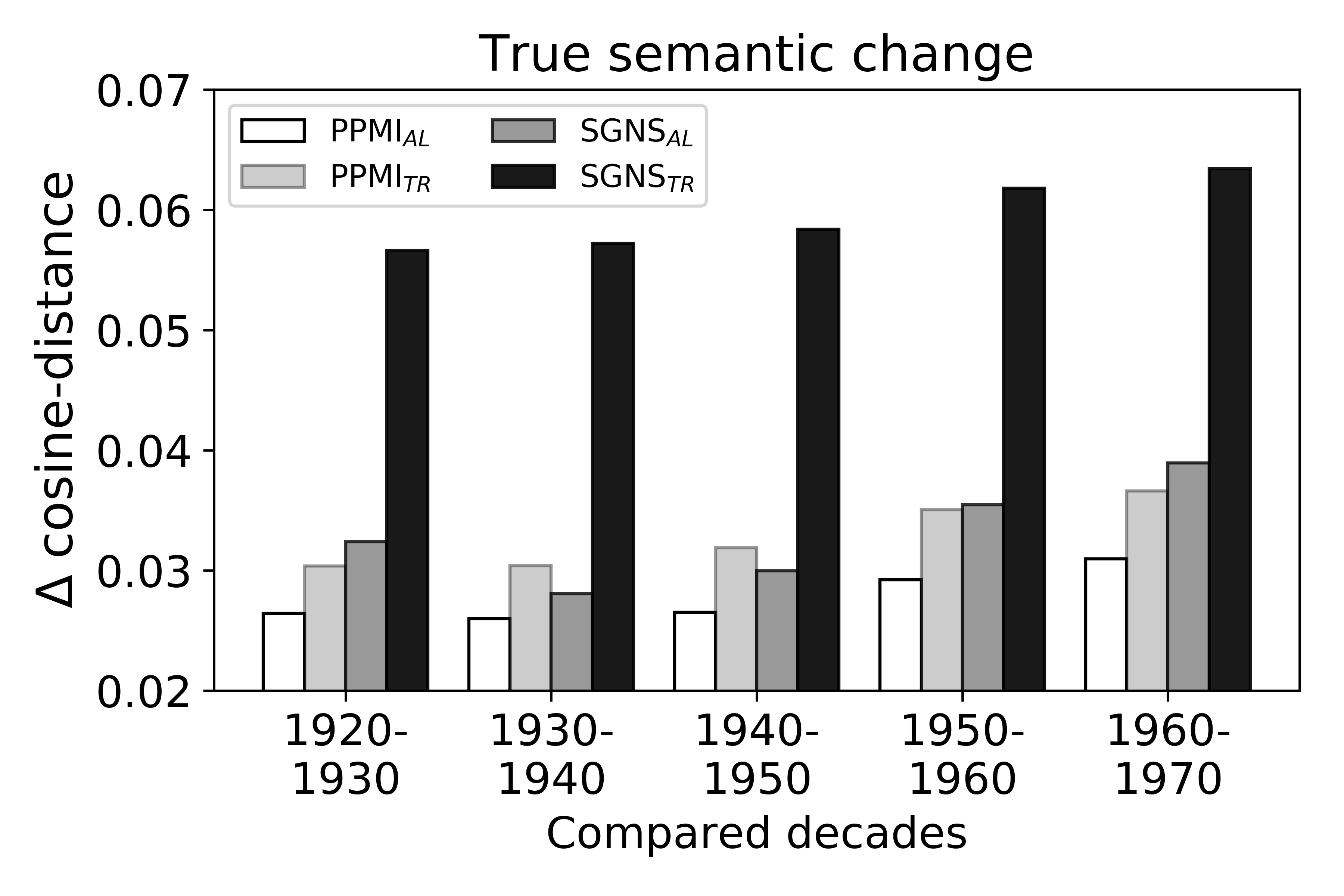}}
\caption{\label{fig:my-label} Comparison of aligned embedding spaces and temporal referencing using both the genuine and the shuffled corpora. High difference in cosine distance indicate less noise captured by the model. }
\end{figure}

We start analyzing the true semantic change for each of the models (PPMI$_{\text{AL}}$ to PPMI$_{\text{TR}}$ and SGNS$_{\text{AL}}$ to SGNS$_{\text{TR}}$) over the corpus. In Figure \ref{fig:my-label}, we can see that temporal referencing introduces less noise throughout the 5 decadal comparisons. For both PPMI and SGNS, the true semantic change increases for the TR models compared to the aligned. 

Importantly, Table \ref{table:avgcosgrouped} shows that for the PPMI models, Temporal Referencing has a much smaller improvement over the aligned model (.005) compared to the SGNS models (.026) (all reported differences are statistically significant, t-test $p<.01$). Temporal Referencing influences the PPMI models only by creating more stable context vectors. In contrast, for the SGNS models the introduction of Temporal Referencing circumvents the use of alignment in addition to creating more stable context vectors. Therefore, the results support our hypothesis that TR has two complementing factors that improve prior models; firstly, it avoids the need for alignment altogether (and the noise that usually comes with it), and secondly, it produces more stable context vectors due to the increased volume of data when using the full corpus.  

\begin{table}[h]
\caption{Difference in average cosine distance between genuine and shuffled conditions (true semantic change) for each method, collapsed over the 5 time bins (1920--1970) in COHA.}
\label{table:avgcosgrouped}
\centering
\begin{tabular}{lllc}
\toprule
& Align & TR & $\Delta$\\
\cmidrule(l){2-4} 
SGNS& 0.033& 0.059 & 0.026\\
PPMI &0.028&0.033 & 0.005\\
\bottomrule
\end{tabular}
\end{table}

\paragraph{Smoothness of Temporal Referencing} We further analyzed the nature of the progression of the cumulative semantic change that words exhibit over time. Under the assumption that words change their meaning in a systematic way, it follows that words' semantic change would increase over the years. 
Therefore, an ecologically valid model of semantic change should show that the words change more as the time interval for comparison increases, for the vocabulary as a whole. In contrast, if a model captures stochastic fluctuations in the words' vectors instead of true semantic change, then such a shift in the distribution will be less prominent.

\begin{figure}[t]
\center{\includegraphics[scale=0.4]        {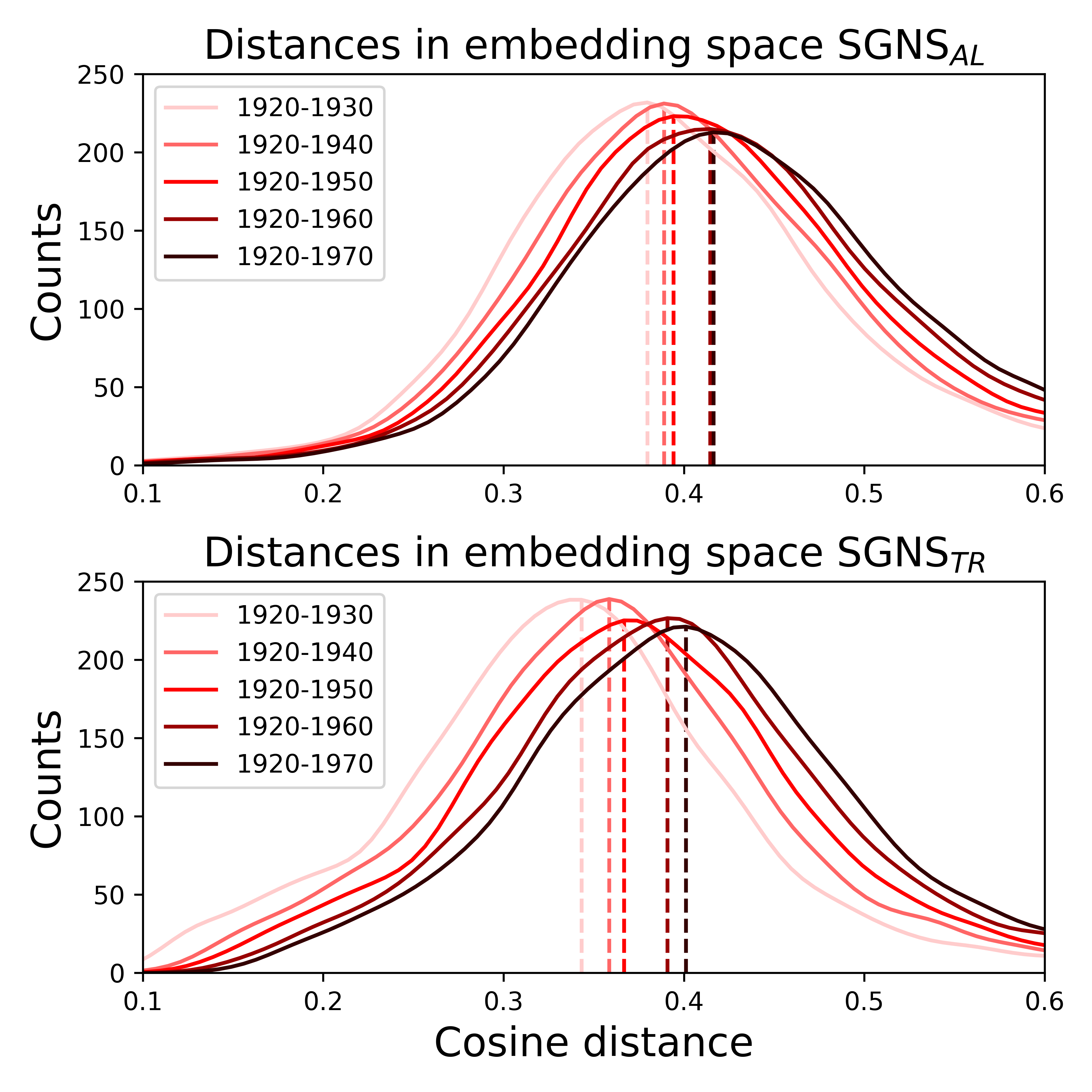}}
\caption{\label{fig:histograml} Smoothed histograms of word distances for the two SGNS models. For the TR model, we see a more constant cumulative shift which is reflected by the overlap between the distributions as well as by differences in their means (dashed vertical lines).}
\end{figure}

We plot the distribution of the words' cosine distances with increasing time intervals (relative to 1920) for both SGNS models in Figure~\ref{fig:histograml}. Both models show a gradual transition from left (smaller change scores) to right (larger change scores). This corroborates our basic assumption that words change more as the time interval for comparison increases. Crucially, Temporal Referencing shows a more constant cumulative progression of cosine distances over time in contrast to alignment where decadal cosine distance distributions seem to be more volatile. We follow \citet{bamler17} in interpreting these results as attesting for the relatively high noise factor in the SGNS$_{\text{AL}}$ over the SGNS$_{\text{TR}}$. 

Overall, the different analyses converge to the same conclusion: Temporal Referencing is a better model for capturing a word's semantic information from diachronic text because it introduces less noise. Next, we will investigate if a less noisy model is also better at detecting semantic change.

\subsection{Experiment 2: Synthetic semantic change}
\label{subsec:experiment2}

This experiment aims to see how well our methods can find different synthetic change types. 
In order to minimize natural semantic change in the dataset, we made use of the synchronic dataset COCA which we randomly shuffled, and simulated a diachronic corpus for which we have 7 time-bins. 
We randomly assigned a seventh of COCA to each of our artificial time periods, labeled $t_1$ to $t_{7}$. Sentences in which either word of the synthetic semantic change pairs (see Sec. \ref{sec:SynLSC}) or their corresponding control words appeared were held out. These sentences were subsequently added back to COCA according to the procedure outlined in Section \ref{sec:SynLSC}, which enabled us to control for the fixed ratio incremental steps between the recipient and donor words (i.e., changes to the injection ratio were made only for $t_2$-$t_3$, $t_3$-$t_4$, $t_4$-$t_5$, and $t_5$-$t_6$, while $t_1$-$t_2$ and $t_6$-$t_7$ had no such changes).

All four models were trained on the 7 synthetic time-bins exactly as in Experiment 1. The target words were the 356 words with synthetic lexical semantic change and their 356 control words that were matched with the same frequency increase but otherwise are considered semantically stable. For each target word, the cosine distances between two consecutive synthetic time-bins were computed, resulting in 6 change scores per word.

We analyze the peak distribution of the individual words. We defined the peak position of each word as its vector $argmax$ (the position in which it shows the maximum cosine distance). 
In order to evaluate the models' ability to truly detect semantic change, we formulate a naïve binary classification task based on the words' peak positions. For each word, if the peak is in position 2--5, we classify it as changed, and otherwise as stable and measure accuracy and F1-score. 

\begin{figure}[h]
    \center{\includegraphics[width=1.0\linewidth]   {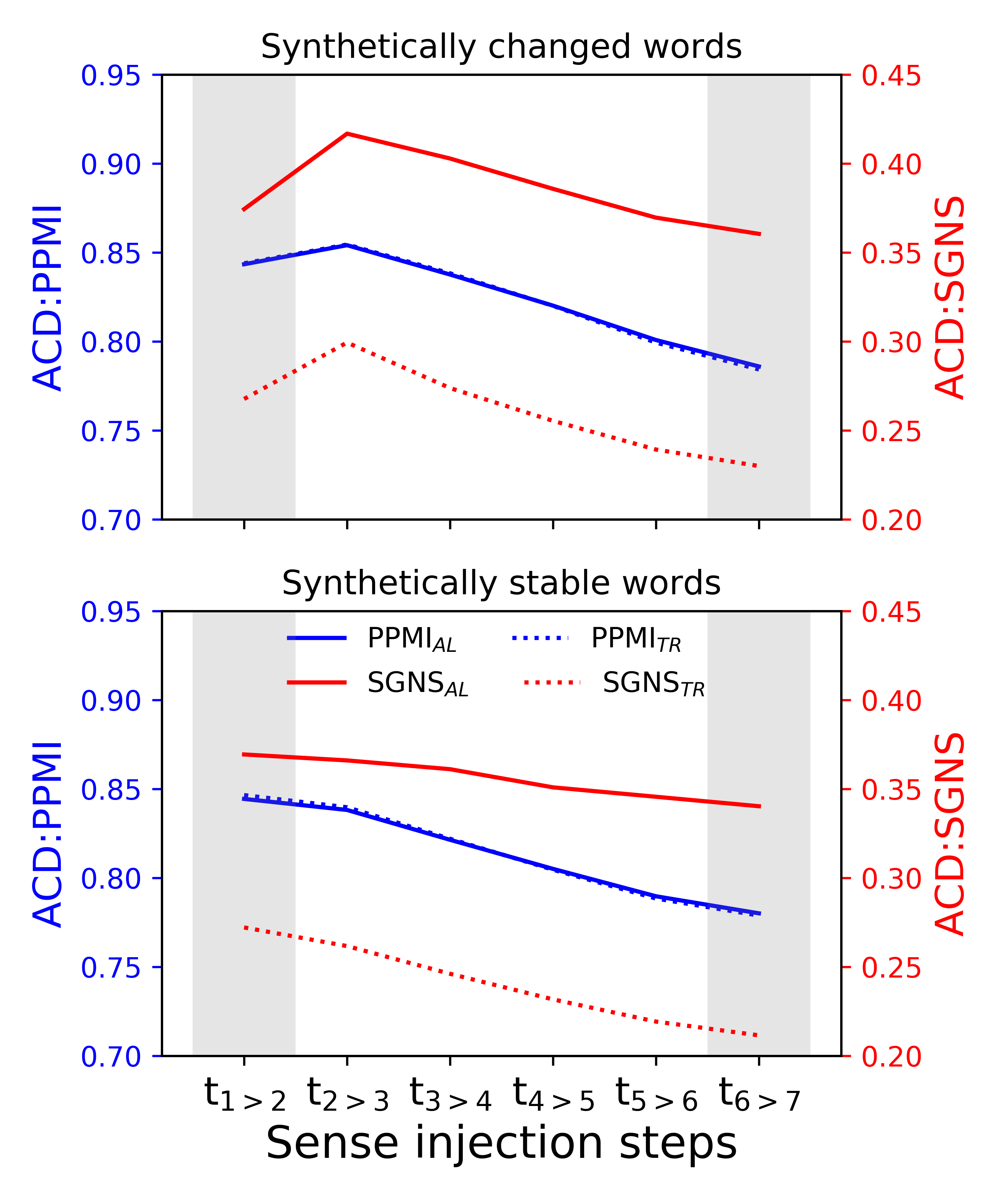}}
    \caption{\textit{acd} at different sense injection steps for the four models. Steps without sense injection are shaded. \label{fig:avgSC} }
  \end{figure}

\paragraph{Results} Figure \ref{fig:avgSC} shows the \textit{acd} of the four models for the change and stable words separately, according to the different sense injection ratios. The two plots differ markedly. For the semantic change words (upper plot), all four models show a noticeable peak when the new sense was first injected (step 2), followed by a steady decrease in \textit{acd} until step 6. In contrast, the stable words only show the steady decrease starting from step 1, without any noticeable peaks. This decrease probably stems from the target words' increased frequency that can lead to more accurate word embeddings \cite{HellrichH16-Coling}. Because peaks in \textit{acd} are interpreted as points were semantic change was the most profound, the results support the models' ability to detect synthetic semantic changes. 

Although the majority of peaks for the semantic change words fall in step 2, as expected by the \textit{acd} analysis above, words had their peaks in other step positions as well (see Appendix~\ref{sec:peak_analysis}).\footnote{We also ran experiments with moving the time point when the first change was injected and the results mimic those presented here.}

Table \ref{tab:exp2accuracy} reports accuracy and F-scores for the four models in the binary classification task. As clearly seen, all four models perform better than chance even under these very rudimentary conditions (finding the $argmax$ of a vector of length 6). Crucially, SGNS$_{\text{TR}}$ outperforms the rest of the models, and especially SGNS$_{\text{AL}}$ that shows the worst performance. These results corroborate our hypothesis from Experiment 1 that noise is negatively influencing task performance. By alleviating the noise factor that exists in SGNS$_{\text{AL}}$ (due to alignment), SGNS$_{\text{TR}}$ is able to show substantial gains in this binary classification task. 

\begin{table}[h]
    \centering
    \small
      \caption{Accuracy (averaged, and split into individual classes) and F1-scores for semantic change detection. For stable words (control words), peaks at 1 and 6 steps are correct. For change words, peaks at steps 2--5 are correct. We see that all methods find unrelated change better than related change, and that SGNS$_{\text{TR}}$ outperforms the other methods.}
    \label{tab:exp2accuracy}
    \begin{tabular}{l l l l l@{}}
    \toprule
    & PPMI$_{\text{AL}}$ & 	PPMI$_{\text{TR}}$	& SGNS$_{\text{AL}}$ & SGNS$_{\text{TR}}$\\
    \cmidrule{2-5}
Stable  &	0.52&	0.54&	0.37	&\textbf{0.57}\\
Unrelated	&0.83&	0.83&	0.86&	\textbf{0.91}\\
Related&	0.73&	0.73&	0.78&	\textbf{0.78}\\
\midrule
 Mean acc.&	0.65&	0.66&	0.59&	\textbf{0.70}\\
F1-score & 0.69& 0.69 & 0.67& \textbf{0.74}\\
\bottomrule
    \end{tabular}
\end{table}

\begin{figure}[t]
\center{\includegraphics[scale=0.6]{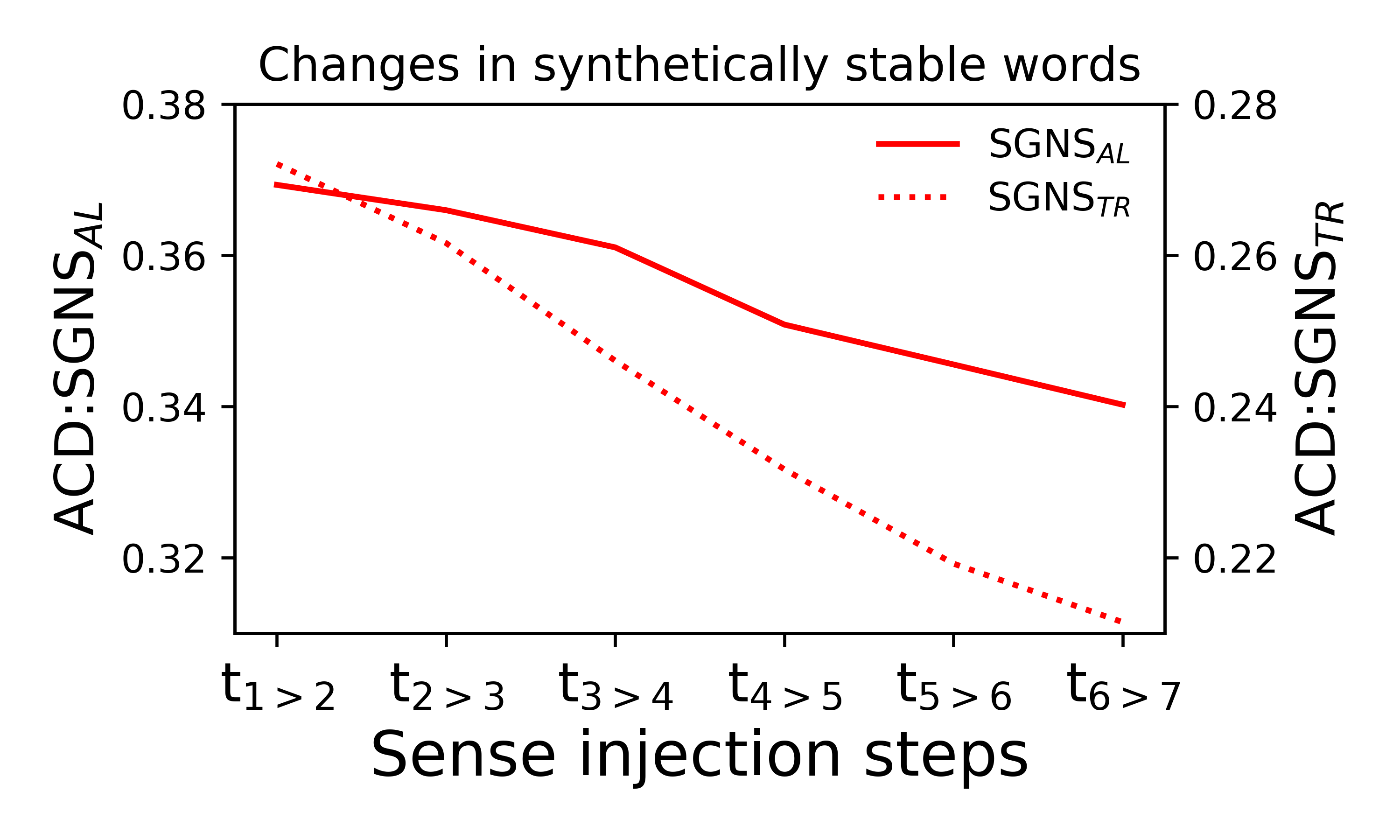}}
\caption{\label{fig:dominik} Mean cosine distance curves for SGNS$_{\text{TR}}$ and SGNS$_{\text{AL}}$.}
\end{figure}

\paragraph{Discussion} Table \ref{tab:exp2accuracy} shows that SGNS$_{\text{TR}}$ gains its performance advantage over SGNS$_{\text{AL}}$ mainly from a better classification of the stable words ($0.37$ vs. $0.57$). In order to understand this better, we inspect their mean cosine distance curves only for stable words in Figure \ref{fig:dominik}. SGNS$_{\text{TR}}$'s curve clearly declines, while SGNS$_{\text{AL}}$'s curve declines much less and is more volatile. We attribute the decline of both curves to the diminishing noise that comes from the continuous increase in frequency of the control words \citep{Dubossarsky-EMNLP17}. It seems that this diminishing frequency noise is counteracted by the alignment noise, yielding a flatter curve for SGNS$_{\text{AL}}$. The latter increases SGNS$_{\text{AL}}$'s chance to have peaks in one of the center injection steps producing false positives in our classification task. However, this property may also have a positive influence on SGNS$_{\text{AL}}$ in related LSC detection tasks \citep{Schlechtwegetal19}.

\subsection{Experiment 3: WSC testset}

So far, the results have been based on either a large random sample to show general tendencies for the language in the corpus as a whole, or synthetically injected semantic change.  In this part, we test the behavior of our methods on a small, manually created testset for semantic change. We use the Word Sense Change Testset \cite{WSCtestset} that consists of words and the different associated change events, for the time span 1785 -- 2010. In this experiment, we ignore the sense changes and consider only words as changed or stable, and restrict our change words to those that have change events between 1920 and 1970.\footnote{As an example, the word \textit{car} is considered stable since its change event occurred before 1920. }
In total we have 13 changed and 19 stable words (excluding words with a total frequency $\leq 100$). 

\begin{table}[h]
\caption{\textit{acd} for WSC testset. Var $\in (0.0-0.01)$.  CH = changed word, ST = stable word, DIFF = difference between ACD for change and stable in percent.  \label{tab:ACD-WSC}}
\centering
\small
\begin{tabular}{lllll}
\toprule
&\multicolumn{2}{c}{SGNS} & \multicolumn{2}{c}{PPMI}\\
& Align & TR & Align & TR\\
\cmidrule(l){2-3}
\cmidrule(l){4-5}
CH &0.47  & 0.31   & 0.86  & 0.86 \\
ST & 0.34   &0.21 & 0.71  &  0.73 \\
\midrule
DIFF & 38\% & 50\% & 20\% & 17\%\\
\bottomrule
\end{tabular}
\end{table}

In Table \ref{tab:ACD-WSC} we see \textit{acd} of each model on the changed and stable words. 
We find that for all methods, SGNS$_{\text{AL}}$, SGNS$_{\text{TR}}$, PPMI$_{\text{AL}}$ and PPMI$_{\text{TR}}$, the \textit{acd} for the changed words is statistically significantly higher (p values $\leq 0.01$) than for the stable words which nicely corresponds to intuition; words with true semantic change should have vectors that differ more than words without change.
The mean difference between the stable and the changed words, that gives us some notion of how well the two different classes are separated, is highest for SGNS$_{\text{TR}}$. Because of the limited size of the testset, the results are indicative rather than conclusive and we continue with a manual analysis of the nearest neighboring words.

\begin{figure}[t]
       \center{\includegraphics[width=1.0\linewidth]  {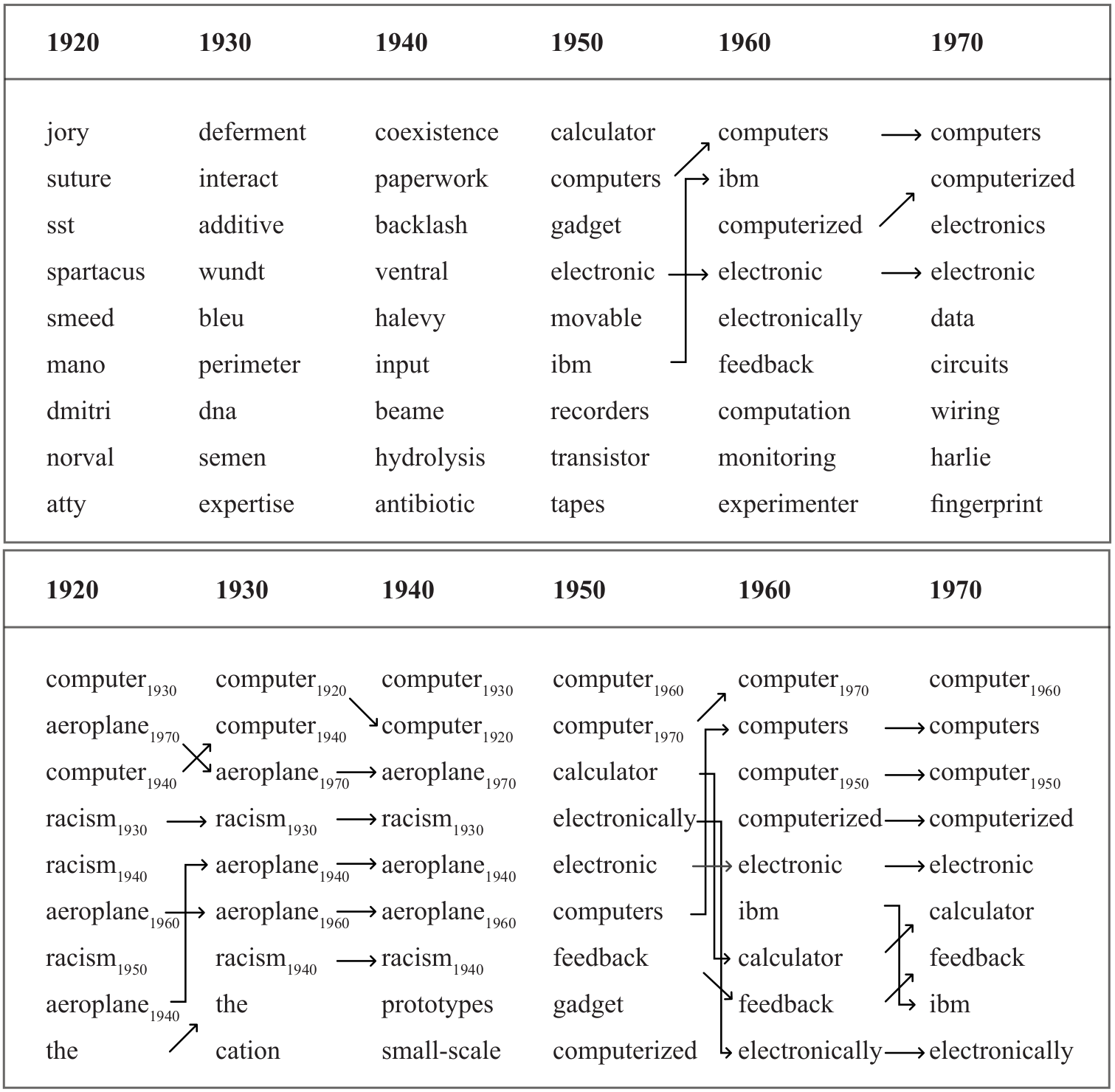}}
      \caption{\label{fig:knn-comp-paper}Nearest neighbors for \textit{computer}. Upper part SGNS$_{\text{AL}}$, lower part SGNS$_{\text{TR}}$. A larger rendering of this figure is available in Appendix~\ref{sec:appComp}.}
     \end{figure}

We carry out a qualitative evaluation for the closest neighbors for \textit{computer} (see Figure~\ref{fig:knn-comp-paper}), a word we expect to have changed after the invention of the digital computer in the 1940s, for the SGNS aligned version and SGNS with Temporal Referencing. 
SGNS$_{\text{AL}}$ has only a few words in common in 1950--1970, and while the digital computer is showing here, there are few overlapping words. The time periods 1920--1940 have no common words.  In comparison, the SGNS$_{\text{TR}}$ show clear patterns. We see a clear break between 1940 and 1950, without any overlapping word, and a pattern between 1950--1970;  the closest words are the other \textit{computer$_{\textit{1940--1970}}$}.\footnote{The closest words in 1920--1940 have high cosine distances and are thus not very related. Still, for each \textit{computer$_{\textit{time}}$}, the other vectors of \textit{computer} are among the neighbors, meaning that despite sparsity and little overlap in context,  some structure is found.}  This is exactly the pattern that we expected to see using the sense injection; stable senses can be distinguished from changing senses by their relationship to the other temporally referenced vectors. 

Next, we study a word for which we expect no sense change, namely \textit{ship} (see Appendix~\ref{sec:appShip}).
The SGNS$_{\text{AL}}$ show a fairly low \textit{acd}, but still there are large differences in the top neighboring words. The SGNS$_{\text{TR}}$ show what we expect; the most similar words are the other \textit{ship$_{\textit{1920--1970}}$}, and over time we see that the `self-similarity' decreases. For almost all decades, the most similar words are \textit{ship} from the decade before and after. The lower words also help describe the meaning of \textit{ship}, as a \textit{boat} and later also as a \textit{spaceship}. The pattern of stability is much more clear for SGNS$_{\text{TR}}$ than SGNS$_{\text{AL}}$ and holds for most other stable words as well.  

For the word \textit{tape}, that has a change in dominant sense (or an addition of another strong sense) with the addition of the music tape to adhesive tape, we see the same patterns as for ship, but the bottom words contain \textit{ribbon, paper, adhesive} for 1920--1940 and \textit{recorder, recording, stereo} in 1950--1970.\footnote{Find all nearest neighbour lists at \url{https://github.com/Garrafao/TemporalReferencing/tree/master/data}.}
      
For both the real change in
Table~\ref{tab:ACD-WSC} and the  synthetic  change in Table \ref{tab:ACD-synth}, we find that SGNS$_{\text{TR}}$ is best at differentiating between the stable and the change classes for both datasets (50\% for WSC and 26\% for synthetic change).

\begin{table}[t]
\caption{\textit{acd} for synthetic change. Var $\in (0.0-0.01)$.   \label{tab:ACD-synth}}
\centering
\small
\begin{tabular}{lllll}
\toprule
&\multicolumn{2}{c}{SGNS} & \multicolumn{2}{c}{PPMI}\\
& Align & TR & Align & TR\\
\cmidrule(l){2-3}
\cmidrule(l){4-5}
CH &0.46  & 0.33  & 0.86  & 0.87 \\
ST & 0.37   &0.26 & 0.83  &  0.83 \\
\midrule
DIFF & 24\% & 26\% & 4\% & 4\%\\
\bottomrule
\end{tabular}
\end{table}

\section{Conclusions and future work}
In this paper, we have empirically tested the temporal referencing method for lexical semantic change. We train one vector space model over the whole corpus, and thus share information of the context words while training individual vectors for each target word and time period. We compare two commonly used models, namely PPMI and SGNS because of their properties; the PPMI model is count-based and does not require alignment across time, while the SGNS model has shown state-of-the-art results in previous work. 

We find that the SGNS model trained with Temporal Referencing contains significantly less noise than the standard SGNS for which an alignment is necessary. In comparison, for the PPMI model where no alignment is needed, Temporal Referencing also significantly reduced the noise level, but to a lesser extent. 

Next we evaluated whether the noise reduction carries over performance on a synthetic lexical semantic change detection task. We simulated change in a controlled and semantically principled way, using sense injection and showed that words with semantically related and unrelated semantic change can be differentiated from control (stable) words that are not sense injected, but increase in frequency in the same way as the changed words. SGNS with Temporal Referencing outperforms the other methods in correctly classifying the words to the two classes (change vs. stable). 

Finally, we evaluated on a small, handcrafted set of change and stable words and found that SGNS with Temporal Referencing gives the largest separation between words that undergo semantic change and those that stay stable over time. In particular, we observe a similar behavior between this smaller testset and the synthetic sense injection, supporting our sense injection method as a good proxy for isolating and studying lexical semantic change. 

Our results support the following conclusion; \textit{trained on a diachronic corpus, SGNS with Temporal Referencing will capture more true semantic change.} 
In the future, we plan to evaluate Temporal Referencing against the related dynamic embedding models on an annotated empirical lexical change dataset with multiple languages. We also plan on testing how well Temporal Referencing deals with corpora that are too small for alignment-based methods, hopefully opening new avenues of quantitative research. 

\section*{Acknowledgements}
The authors would like to thank Dr. Barbara McGillivray for her encouragement, and the anonymous reviewers for their helpful comments and suggestions. 
This work has been funded in parts by the University of Helsinki (research visit grant C1/2019 from the Faculty of Arts, to SH), by the project \textit{Towards Computational Lexical Semantic Change Detection} supported by a project grant (2019–2022; dnr 2018-01184, to NT), the Centre for Digital Humanities at University of Gothenburg, the Konrad Adenauer Foundation and the CRETA center funded by the German Ministry for Education and Research (BMBF), and the Blavatnik Postdoctoral Fellowship.

\bibliography{bib}
\bibliographystyle{acl_natbib}

\clearpage
\appendix

\section{Pre-processing and Hyperparameter Details}
\label{sec:parameter}

We lower-cased all tokens in the corpora before extracting word-context pairs. For pair extraction we chose a window size of 5 for both, AL and TR. Corpus tokens were skipped as word or context if they did not have a minimum frequency of 100 in the full corpus used (i.e., 1920-1970 for COHA and full COCA) or contained non-alphabetic characters (except hyphens).

We tuned model parameters on the most recent time bin of COHA (2000-2009) based on word similarity task scores \citep{hill2015simlex,Finkelstein:2001} reaching near state-of-the-art results \citep{Levy2015}. The parameters for SGNS were $dim=300$ (vector dimensionality), $cds=0.75$ (context distribution smoothing), $k=5$ (number of negative samples) and $ep=1$ (number of training epochs). PPMI was smoothed and shifted \citet{Levy2015}. The parameters were $cds=0.75$ and $k=5$ (shifting parameter).

\section{Peak distribution analysis}
\label{sec:peak_analysis}

\begin{figure}[!ht]
    \center{\includegraphics[width=1.0\linewidth]   {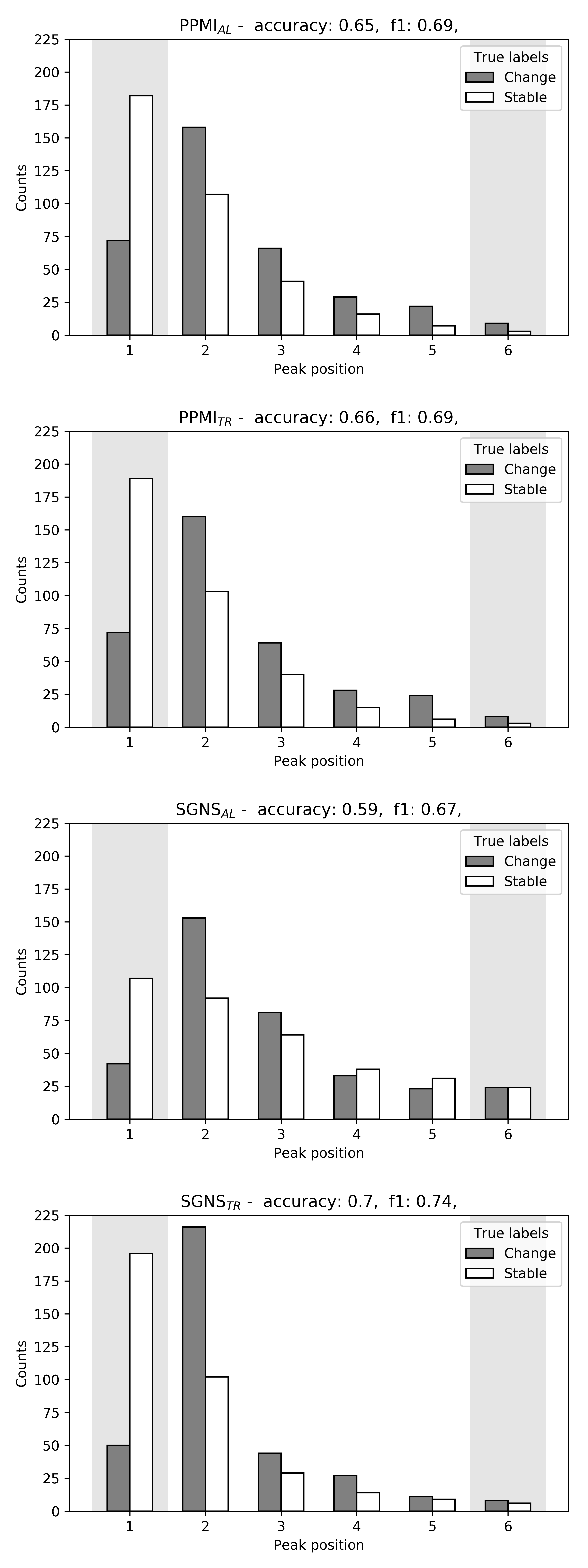}}
    \caption{\label{fig:peakdist} Distributions of peak detection for our four models for synthetically changed (grey) and stable-control (white) words. Accuracy and F-scores are reported for each model at the top of the panels. Shaded areas represent steps with no sense injection.}
  \end{figure}

In Figure \ref{fig:peakdist} we present the peak distributions of the four models for the 712 target words (356 changed and 356 stable), color coded according to the true classification (change/stable). The peaks represent the models' predictions with respect to where the maximal cosine distance is found for each word, which we later use in a naive and rudimentary binary classification task.
As can be seen from the different distributions, all models frequently find peaks in position 2 (corresponding to the event of the first sense injection). However, they are still very much different in their overall peak distributions which influence their sensitivity in detecting synthetically semantic changed words (Table \ref{tab:exp2accuracy}).

\section{WSC TestSet}
  
 In Table~\ref{tab:changedWSC} we list the words that have undergone semantic change, as well as the change year(s) and a description of the change. In Table~\ref{tab:stableWSC} we list words that do not have changed meanings.

\begin{table*}[htbp]
\caption{Changed words from WSC Testset}
\label{tab:changedWSC}
\centering
\begin{tabular}{lll}
\toprule
 \textbf{Word}& \textbf{Change year} & {Description}\\
\midrule
aeroplane & 1919-1920 & First use as weapon of war and commercial flights\\
cinema& ~1900 & movie theatre\\
computer& ~1940 &  digital computer\\
cool& ~1964 & a way of being\\
flight& ~1918 & after WWI commercial aviation grows rapidly\\
gay& 1985 &  recommended for use instead of homosexual\\
memory& 1960 & digital memory\\
mouse& 1965 & the computer mouse was introduced\\
record& 1920 & electrical music records\\
rock& 1950-1960 & birth of rock music\\
tank& 1917 & first tank in battle\\
tape& 1960 & common household use of the magnetic tape\\
\bottomrule
\end{tabular}
\end{table*}

\begin{table}[htbp]
\caption{Stable words from WSC Testset}
\label{tab:stableWSC}
\centering
\begin{tabular}{ll}
\toprule
automobile & music \\
bank & newspaper \\
camera & paper \\
car& phone \\
deer& ship \\
export& symptom \\
founder& telephone \\
horse& train\\
mail& travel\\
mirror & \\
\bottomrule
\end{tabular}
\end{table}

  \section{Closest Neighbors for \textit{Computer}}\label{sec:appComp}
  In Figure \ref{fig:knn-computer} we see the closest neighbors for \textit{computer}, a word we expect to have changed after the invention of the digital computer in the 1940s, for the SGNS aligned version (upper) and SGNS with temporal referencing (lower). 
  
   \section{Closest Neighbors for \textit{Ship}}\label{sec:appShip}
\begin{figure*}[t]
        \center{\includegraphics[width=.8\linewidth]   {computer-NN.png}}
        \caption{\label{fig:knn-computer}Nearest neighbors for \textit{computer}. Upper part SGNS$_{\text{AL}}$, lower part SGNS$_{\text{TR}}$. }
           \center{\includegraphics[width=0.8\linewidth]  {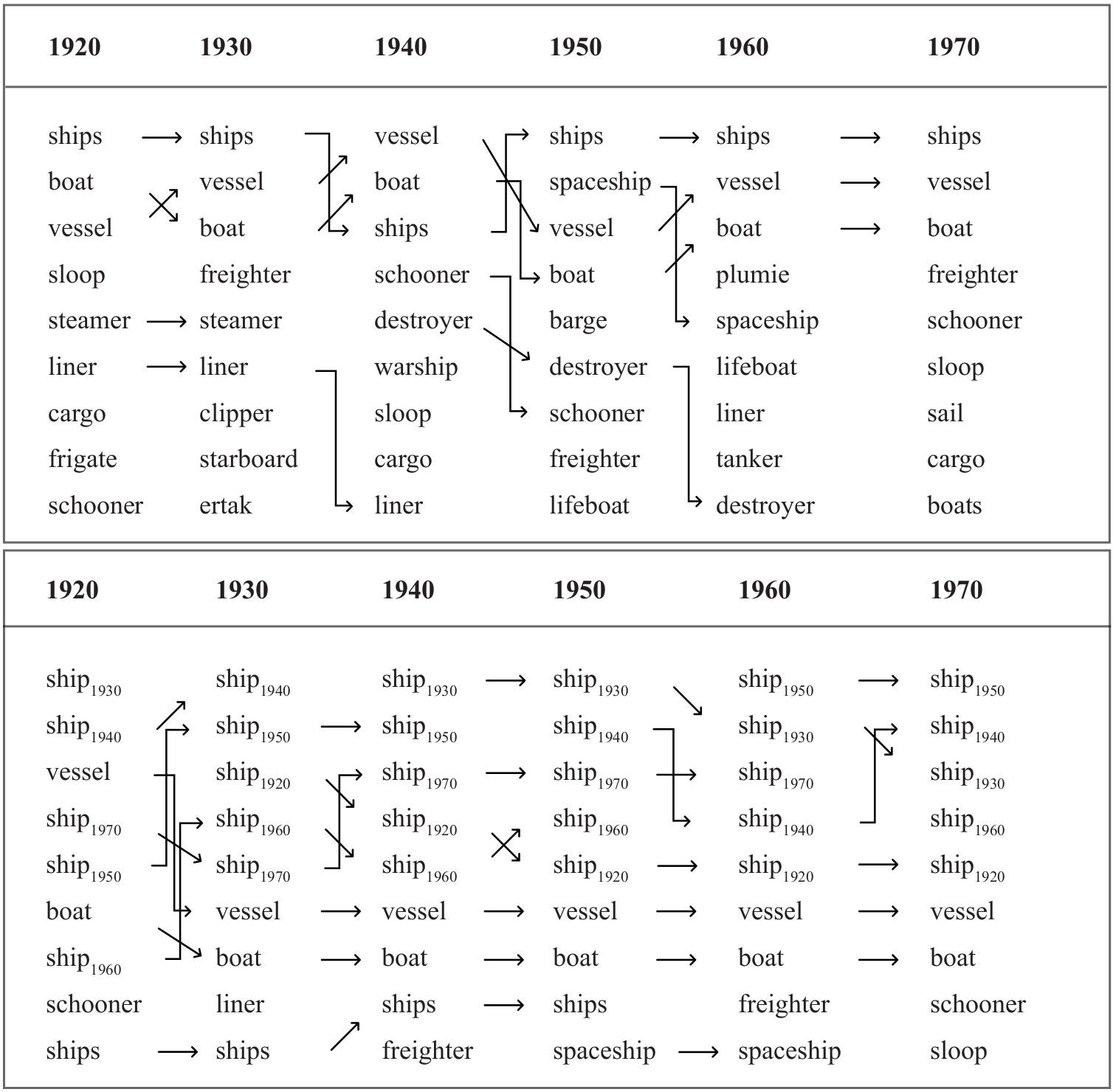}}
      \caption{\label{fig:knn-ship}Nearest neighbors for \textit{ship}. Upper part SGNS$_{\text{AL}}$, lower part SGNS$_{\text{TR}}$. }
      \end{figure*}

In Figure \ref{fig:knn-ship} we see the closest neighbors for \textit{ship}, a word we expect to be stable, for the SGNS aligned version (upper) and SGNS with temporal referencing (lower).

\end{document}